\documentclass[letterpaper, 10 pt, conference]{ieeeconf}  % Comment this line out if you need a4paper

\IEEEoverridecommandlockouts                              % This command is only needed if 
\usepackage{comment}
\usepackage{xcolor}
\usepackage{graphicx}
\usepackage{booktabs}
\usepackage{flushend}
\usepackage{biblatex} %Imports biblatex package
\title{\LARGE \bf
Enhancing Robot Explanation Capabilities through Vision-Language Models: a Preliminary Study by Interpreting Visual Inputs for Improved Human-Robot Interaction%*%*
}

\author{David Sobrín-Hidalgo$^{*,1}$, Miguel Á. González-Santamarta$^{1}$, Ángel Manuel Guerrero-Higueras$^{1}$,\\Francisco J. Rodríguez-Lera$^{1}$, and Vicente Matellán-Olivera$^{1}$% <-this % stops a space
\thanks{$^{*}${Correspondence to: \tt\small dsobh@unileon.es}}%
\thanks{$^{1}$Robotics Group, University of León, Campus de Vegazana s/n, 24071 León, Spain.}%
}

\begin{document}

\maketitle
\thispagestyle{empty}
\pagestyle{empty}

%%%%%%%%%%%%%%%%%%%%%%%%%%%%%%%%%%%%%%%%%%%%%%%%%%%%%%%%%%%%%%%%%%%%%%%%%%%%%%%%
\begin{abstract}

This paper presents an improved system based on our prior work, designed to create explanations for autonomous robot actions during Human-Robot Interaction (HRI). Previously, we developed a system that used Large Language Models (LLMs) to interpret logs and produce natural language explanations. In this study, we expand our approach by incorporating Vision-Language Models (VLMs), enabling the system to analyze textual logs with the added context of visual input. This method allows for generating explanations that combine data from the robot's logs and the images it captures. We tested this enhanced system on a basic navigation task where the robot needs to avoid a human obstacle. The findings from this preliminary study indicate that adding visual interpretation improves our system's explanations by precisely identifying obstacles and increasing the accuracy of the explanations provided.

\end{abstract}

%%%%%%%%%%%%%%%%%%%%%%%%%%%%%%%%%%%%%%%%%%%%%%%%%%%%%%%%%%%%%%%%%%%%%%%%%%%%%%%%
\section{INTRODUCTION}

The merging of Human-Robot Interaction (HRI) and eXplainable Artificial Intelligence (XAI) has led to the development of eXplainable Autonomous Robots (XAR), considered key for enhancing trust and comprehension in autonomous robots by humans \cite{trustworthy_explainability}. This paper extends our previous research, ``Explaining Autonomy: Enhancing Human-Robot Interaction through Explanation Generation with Large Language Models''~\cite{sobrínhidalgo2024explaining}. In our initial work, we presented a system that generates understandable explanations for the behavior of autonomous robots using Large Language Models (LLMs) \cite{LLMsSurvey} and prompt engineering \cite{sahoo2024systematic}, specifically a Retrieval Augmented Generation (RAG) method \cite{lewisRAG}, focusing on making the robots' actions clear and trustworthy to users by analyzing the logs these systems produce.

In HRI, the reliability and trustworthiness of robot behavior are essential \cite{trustworthy2015}. Consequently, there is ongoing research into methods that can increase human trust in robots. Explainability aims to make the robots' decision-making and behavior more transparent and understandable to humans. Our earlier efforts in this area focused on explaining robot behavior through log analysis and designing a system that provides explanations based on the logs created during a robot's operation.

However, relying just on textual data can limit the accuracy of explanations in environments where visual context plays a significant role in robot actions and decisions. Acknowledging this, we have expanded our approach to include Vision-Language Models (VLMs) \cite{VLMSurvey}, using inputs from the robot's onboard camera. This edition seeks to enhance the explanations with greater context awareness and a reflection of the environment in which the robot operates.

Our integration of VLMs introduces a novel method for improving explainability in robotics. Although previous research in XAR and XAI has typically focused on either textual or visual explanations independently (as reviewed in \cite{XAIReview} and \cite{xarReview}), our project investigates how combining these two types of explanations can produce better outcomes.

Our work draws inspiration from recent advancements in Large Language Models (LLMs) and their application in robotics, as seen in projects using LLMs like ChatGPT to create natural language explanations \cite{gonzalez2023Haiss}. Additionally, we introduce a new approach by incorporating Vision-Language Models (VLMs) to enhance these explanations, allowing us to interpret visual data alongside text.

Previously, we generated satisfactory explanations by analyzing the textual logs from robots. Yet, we noticed a limitation in accuracy, especially when the logs indicated a possible obstacle in the robot's path. This limitation stems from the fact that textual data alone cannot conclusively identify obstacles encountered during navigation. In this paper, we explore how integrating visual information from the robot's camera can address this issue.

Hence, this paper adds to the field of explainability by showcasing how VLMs can enhance autonomous robots' ability to explain their actions. We conduct a small proof of concept to demonstrate that merging textual data from logs with visual inputs can significantly enhance the quality and precision of explanations generated by an LLM for users.

This paper is structured as follows: Section \ref{sc:materials} describes the implementation of our previous work and introduces the expanded system presented in this paper. It also details the experiment conducted to evaluate the system. Section \ref{sc:results} discusses the results obtained from the experiment. Finally, Section \ref{sc:conclusions} points out the conclusions of the work.

\section{MATERIALS AND METHODS}
\label{sc:materials}

% \begin{figure*}[thpb]
% \centering
% \includegraphics[width=0.9\textwidth]{img/RAG_METHOD.png}
% \caption{Original workflow of the explainability system~\cite{sobrínhidalgo2024explaining}.}
% \label{fig:workflowORG}
% \end{figure*}

In this section, we look back at our initial system that uses LLMs to create explanations based on robot logs \cite{sobrínhidalgo2024explaining}. Since we use llama\_ros \cite{llama_ros_2023} to execute LLMs, as in other works \cite{gonzálezsantamarta2023integration}, our system is aimed to work with ROS 2-based robots \cite{macenski2022robot}, specifically the Humble distribution. Next, we introduce a new element to our proposal: a VLM, 
concretely we have chosen the model \textit{cjpais/llava-1.6-mistral-7b-gguf} \cite{liu2023improvedllava}. This addition aims to improve the explanation generation process by analyzing visual data. Finally, we outline the design of our experiment. This experiment aims to test how well this combined explanation system can explain decisions made during a navigation task, especially when the robot encounters an unexpected obstacle. The additional materials used in our experiments include a robot simulator in Gazebo set up within an Amazon Web Services (AWS) environment. 

\subsection{Explanation generation with LLMs}

In our earlier work, we developed a system aimed at enhancing HRI by generating explanations for the behavior of autonomous robots. The foundation of the system is the belief that for humans to trust robots, they must first clearly understand the robots' actions, highlighting the importance of explainability. Thus, the developed system uses an LLM as a log interpreter and explanation generator, alongside an RAG method to refine the LLM context. 

Our system is designed to work with any ROS 2-based robot, regardless of its specific functions, by using the logs produced during the execution. This versatility means the system can be used for different tasks without major changes. The RAG method allows the LLM to use the robot's most relevant logs as context to answer user questions. 

The system works as follows: It starts by collecting the robot's logs, which are then turned into embeddings -- vector representations using an LLM --. We store these embeddings in a vector database, forming a knowledge base. When a user has a question, the system searches this database to find the most relevant information. The LLM then uses these relevant logs, along with the question, to generate an understandable explanation of the robot's actions. %The workflow of the original system is illustrated in Figure \ref{fig:workflowORG}.

This architecture represents a significant advancement towards making autonomous systems more transparent and understandable. By combining the detailed data from textual logs with LLMs' ability to analyze and explain large amounts of data, our system not only helps users gain a better understanding of robot actions but also opens the way for further research into the XAR field.

\subsection{Enhancing explanation capabilities through VLMS}

We've improved our previous proposal by adding a VLM. This new component is designed to understand and interpret the visual information captured by the robot's onboard camera. Our goal with this enhancement is to increase the accuracy of our explanation generation system, which initially depended only on text from robot logs. With the VLM, our system now can create detailed descriptions of the images taken during the robot's tasks, providing a richer context to the LLM that will generate the explanations.

\begin{figure*}[thpb]
\centering
\includegraphics[width=0.9\textwidth]{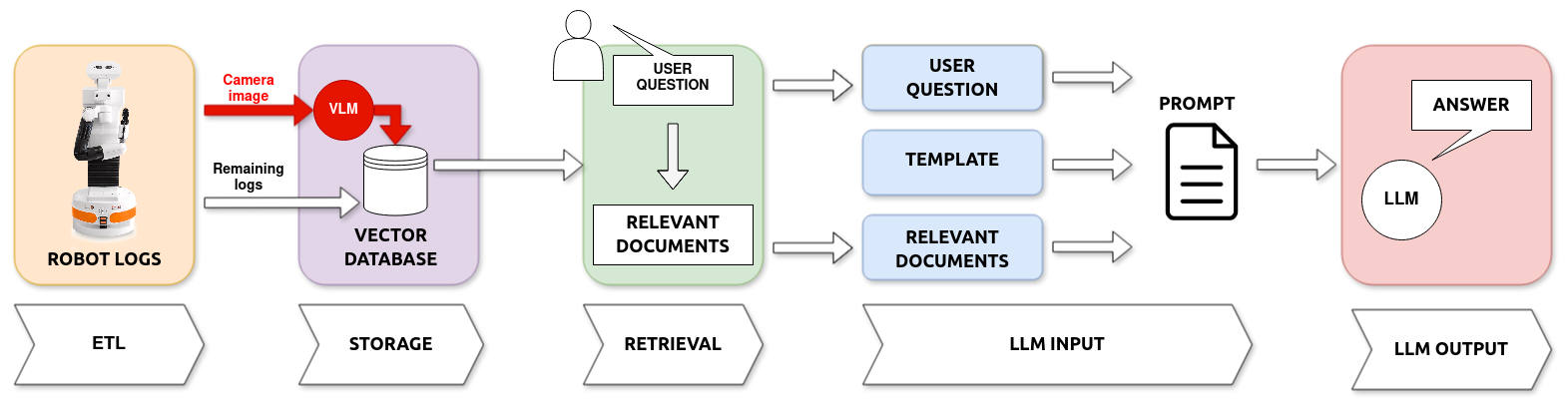}
\caption{New workflow of the explainability system \cite{sobrínhidalgo2024explaining}. The incorporation of the VLM along with the corresponding changes are reflected in the Storage stage.}
\label{fig:workflowNEW}
\end{figure*}

Integrating the VLM involves a complex process where the robot's visual data is first turned into descriptive text that captures the visual scene. This text is then transformed into embeddings, similar to how we process the textual logs. This step helps to link visual understanding with verbal explanations, allowing our system to offer a richer and more accurate description of the robot's environment and actions. The new workflow is illustrated in Figure \ref{fig:workflowNEW}, where the red elements represent the modifications concerning the original system.

To incorporate the VLM capabilities into our previous system, we synchronized the camera topic (\texttt{/camera/image\_raw}) with topics providing relevant navigation information, specifically, the \texttt{/plan} topic. We follow the approach presented in another of our previous works \cite{fernandez2023accountability}. We developed a node that uses the data from this topic to identify potential obstacles indicated by changes in the robot's path. When the node detects a significant increase in the path length, it logs this event in the \texttt{/rosout} topic. Moreover, the VLM processes the image from the camera at that moment. The content of the camera image is converted to text and also logged in \texttt{/rosout}. 

\begin{figure}[thpb]
\centering
\includegraphics[width=\columnwidth]{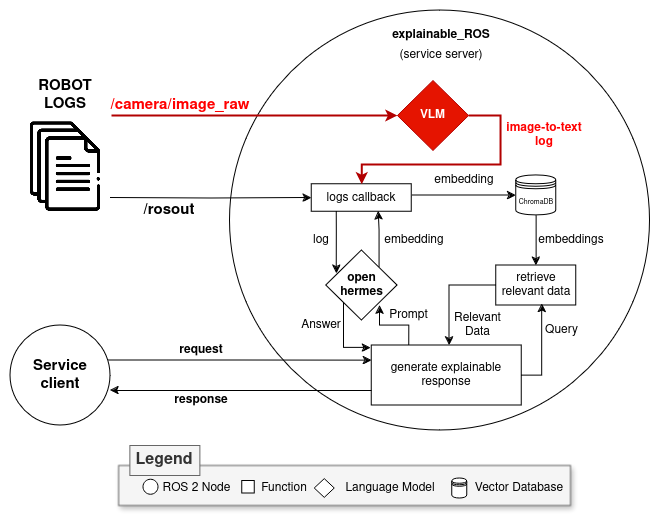}
\caption{New ROS 2-based Explainability System implementation \cite{sobrínhidalgo2024explaining}.
}
\label{fig:implementation}
\end{figure}

Figure \ref{fig:implementation} illustrates the architecture of the new proposal. Elements colored in black represent the original system, while the elements in red illustrate the additions developed for this work. Essentially, both the camera and navigation logs are published to the \textit{/rosout} topic. The explainability node (denoted as \textbf{explainable\_ROS} in the figure) then processes these logs. An LLM function is used to generate embeddings from the processed logs, which are stored in a vector database. When a user inquires about the robot's behavior, the system retrieves the most relevant logs. These logs form the context for the system. This context, along with a predefined template and the user's query, creates the final prompt for the LLM. The model then uses this prompt to generate an explanation for the user through a ROS 2 service.

These new textual descriptions, derived from images, are added to our vector database, enhancing our knowledge base with data from both text and images. This improvement enables the RAG method to draw from a more varied set of data when answering user questions. Consequently, our system can now offer explanations that include the robot's internal workings and actions, as recorded in the logs, and the visual context of its operations.

Adding the VLM is a step forward in our quest for explainable HRI. By merging visual and textual information, our system can generate explanations that are more relevant and accurate. This system can also avoid false positives and true positives related to obstacle detection in the explanation. This advancement in merging LLMs with VLMs to generate explanations is a significant step towards making the behaviors of autonomous robots completely understandable and trustworthy for humans.

\subsection{Evaluation}

To evaluate the improvements made by incorporating a Vision-Language Model (VLM) into our system, we designed an experimental scenario. This test aims to confirm the system's capability to process visual data in real time and use it to provide detailed explanations for autonomous robot behavior. As in our previous work, we have focused the evaluation test on a navigation task inspired by the standard navigation functionality tests of the European Robotics League (ERL).

\paragraph*{Experiment design}

Our experiment takes place in a controlled environment, with the robot required to move from a starting point (Point S) to a goal point (Point G). The experiment was carried out using a simulator of RB1 robot, manufactured by Robotnik. We used a hospital simulation environment made available by Amazon Web Services (AWS)\footnote{https://github.com/aws-robotics/aws-robomaker-hospital-world}.

Initially, the robot's path is free of obstacles, allowing for straightforward route planning and execution. However, at a specific point during the navigation, an obstacle is placed in the robot's path, which leads it to adjust its planned course to avoid it. 

The appearance of the obstacle is a critical moment to test our system's VLM component. When the robot notices a deviation from its initial path, the VLM analyzes the visual data from the robot's camera, identifies the obstacle, and generates a text description of the scene. This dual approach, combining internal navigation log data with external visual information, enriches the context from which our explanations are constructed.

\paragraph*{Evaluation criteria}

After the robot recalculates its path and reaches Point B, it is asked to explain the reason for its route adjustment. Specifically, the question to test our proposal is as follows,

\begin{quote}\it
    ``Pay attention to camera logs. Did the robot encounter any obstacles during navigation? What type of obstacle?''
\end{quote}

This question tests the system's ability to interpret together information from robot logs and camera input. We mainly evaluate the explanation's accuracy, relevance, and clarity. The explanation should accurately recognize the visual context (the presence of the obstacle), detail the decision-making process (the need to reroute), and be understandable and informative to human users.

\paragraph*{Significance}

This experiment not only tests the successful integration of the VLM with our existing system but also highlights the value of interpreting multimodal data to improve the explainability of autonomous systems. By accurately identifying and explaining a route change caused by an unexpected obstacle, our system demonstrates its potential to enhance HRI, where both the physical and operational context influence explanations of autonomous behavior.

\section{RESULTS \& DISCUSSION}
\label{sc:results}

\begin{figure}[thpb]
\centering
\includegraphics[width=\columnwidth]{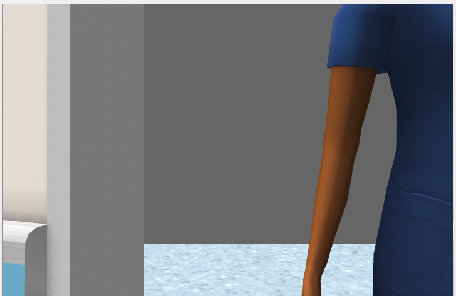}
\caption{Frame evaluated to provide explanation.}
\label{fig:arm}
\end{figure}

\begin{figure}[thpb]
\centering
\includegraphics[width=\columnwidth]{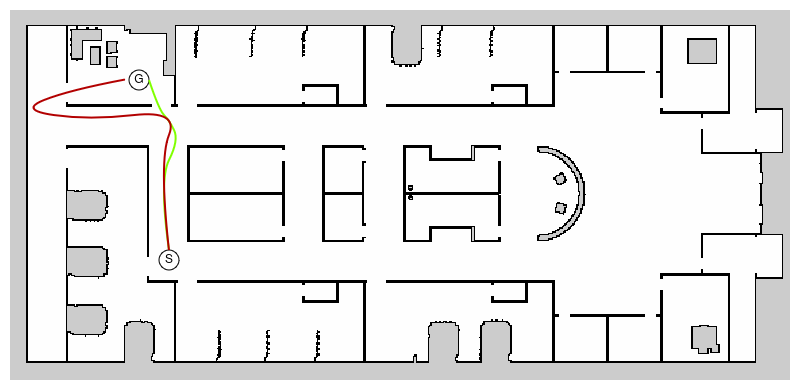}
\caption{Robot trajectories during experiment. S indicates the starting point and G is the goal point. The green line indicates the original trajectories, while the red line indicates the new trajectories after replanning.}
\label{fig:experiment}
\end{figure}

We next analyze the results obtained by our improved explanation system. The key part of this analysis is how the system responds to a query about the obstacles detected during navigation. To facilitate replicability and data transparency, we have made public the rosbag file used during the experiment at \cite{Sobrin_dataset}.

Upon receiving the query: \textbf{\textit{Pay attention to camera logs. Did the robot encounter any obstacles during navigation? What type of obstacle?}}, our system generated a detailed explanation based on the visual data captured by the robot's camera, as well as textual log data with enough verbosity to serve as context. The response was as follows: \textbf{\textit{Based on the provided context with a focus on the camera log, the robot did indeed detect an apparent obstacle while navigating. The mentioned "obstacle" in the image-to-text log describes a person's hand with blue and white stripes. This visual information likely contributed to the navigation system perceiving it as an impediment to its path, resulting in the distance to the goal increasing.}}

This reply shows our system's ability to effectively use camera log visual data, offering a more comprehensive understanding of navigation challenges. An image from the camera, showing a person's arm blocking the robot -- see Figure \ref{fig:arm} --, confirms this explanation, with the arm identified as the navigation issue, highlighting the VLM's precision in identifying visual obstacles. We also visually compare the initial and adjusted navigation routes in Figure \ref{fig:experiment}. This figure illustrates the robot's initial direct path to the goal and the subsequent longer route necessitated by the unexpected obstacle. The deviation underscores the system's capability to dynamically adjust its path in response to real-world challenges, further validating the effectiveness and adaptability of our navigation and explanation systems. The longer route, prompted by obstacle detection, exemplifies the practical application of our system in navigating complex environments and its potential to enhance autonomous decision-making and explanation in HRI.

This outcome illustrates our system's progress in creating explanations informed by both text and visuals. The accurate identification and description of the obstacle -- a person's hand in this instance -- demonstrate our system's potential to improve clarity and understanding in HRI.

\begin{comment}
    
\begin{table*}[ht]
\centering
\small
\caption{User question and LLM's response.}\label{tab:llm_response}
\begin{tabular}{p{.3\textwidth} p{.65\textwidth}}
\toprule
\textbf{User Question} & \textbf{LLM's Response} \\
\midrule
Pay attention to camara logs. Did the robot encountered any obstacle during navigation? What type of obstacle? &
Based on the provided context with a focus on the camera log, the robot did indeed detect an apparent obstacle while navigating. The mentioned "obstacle" in the image-to-text log describes a person's hand with blue and white stripes. This visual information likely contributed to the navigation system perceiving it as an impediment to its path, resulting in the distance to the goal increasing.\\
\bottomrule
\end{tabular}
\end{table*}

\end{comment}

\section{CONCLUSIONS}
\label{sc:conclusions}

Incorporating VLMs into our system goes HRI. This approach builds on the strengths of using Large Language Models to analyze robot logs, now adding visual data to improve explanations. Our tests show that VLMs make explanations more detailed and relevant, particularly when the robot encounters new challenges and needs to change its approach.

The system’s ability to explain navigation decisions, especially around unexpected obstacles, shows the benefits of using both, text and visual data, to create accurate explanations. This is key to making robot actions more transparent and building trust between humans and robots.

Looking forward, integrating VLMs lays the groundwork for future research in XAR, suggesting that incorporating other types of data like audio or tactile feedback could make explanations even richer. Examining how our system can be adapted to various robots and settings is also important.

Our work aims to make interactions between humans and robots more meaningful by enhancing robots’ ability to auto-explain their actions in a human-understandable way, moving us toward a future where robots are more integrated into our daily lives.

\section*{ACKNOWLEDGMENT}

This work has been partially funded by the Recovery, Transformation, and Resilience Plan, financed by the European Union (Next Generation) thanks to the TESCAC  (Traceability and Explainability in Autonomous Systems for improved Cybersecurity) project granted by INCIBE to the University of León; and by the EDMAR (Explainable Decision Making in Autonomous Robots) project, PID2021-126592OB-C21 funded by MCIN/AEI/10.13039/501100011033 and by ERDF ``A way of making Europe''. David Sobrín Hidalgo acknowledges the University of León for its funding support for his doctoral studies.

%%%%%%%%%%%%%%%%%%%%%%%%%%%%%%%%%%%%%%%%%%%%%%%%%%%%%%%%%%%%%%%%%%%%%%%%%%%%%%%%

\printbibliography %Prints bibliography

\end{document}